\documentclass[journal]{IEEEtran}
\usepackage[utf8]{inputenc}
\usepackage[
	final,
	protrusion,
	expansion,
	tracking,
	kerning,
	spacing,
	factor = 1100,
	stretch = 10,
	shrink = 10
]{microtype}
\usepackage{amsmath}
\usepackage{amssymb}
\usepackage{biblatex}
\usepackage{siunitx}
\usepackage[super]{nth}
\usepackage{tikz}
\usepackage{booktabs}

\usetikzlibrary{calc, positioning}
\tikzstyle{image} = [inner sep=0pt]

\newcommand{\true}{\mathcal{T}}
\newcommand{\false}{\mathcal{F}}
\newcommand{\unknown}{\mathcal{U}}

\newcommand{\precision}{\mathbf{P}}
\newcommand{\recall}{\mathbf{R}}
\newcommand{\accuracy}{\mathbf{A}}
\newcommand{\fmeasure}{\mathbf{F}}

\newcommand{\salamander}{\texttt{salamander}}
\newcommand{\fluid}{\texttt{fluid}}
\newcommand{\general}{\texttt{general}}
\newcommand{\kawai}{\texttt{kawai}}
\newcommand{\piano}{\texttt{piano}}
\newcommand{\city}{\texttt{city}}
\newcommand{\shrine}{\texttt{shrine}}

\title{Context-Independent Polyphonic Piano Onset Transcription with an Infinite Training Dataset}
\author{Samuel Li}

\begin{document}
	\markboth{arXiv Preprint}%
{Samuel Li}

	\maketitle

    \begin{abstract}
    	Many of the recent approaches to polyphonic piano note onset transcription require training a machine learning model on a large piano database.
        However, such approaches are limited by dataset availability; additional training data is difficult to produce, and proposed systems often perform poorly on novel recording conditions.
        We propose a method to quickly synthesize arbitrary quantities of training data, avoiding the need for curating large datasets.
        Various aspects of piano note dynamics --- including nonlinearity of note signatures with velocity, different articulations, temporal clustering of onsets, and nonlinear note partial interference --- are modeled to match the characteristics of real pianos.
        Our method also avoids the disentanglement problem, a recently noted issue affecting machine-learning based approaches.
        We train a feed-forward neural network with two hidden layers on our generated training data and achieve both good transcription performance on the large MAPS piano dataset and excellent generalization qualities.
    \end{abstract}

    \begin{IEEEkeywords}
    music information retrieval (MIR), neural networks, data modeling
    \end{IEEEkeywords}

    \section{Introduction}
    \IEEEPARstart{P}{olyphonic} music transcription involves extracting a musical score or equivalent representation from an audio recording.
    In particular, the problem of polyphonic piano onset transcription involves extracting the onset time and pitch of many potentially simultaneous piano notes.
    Deep neural networks have been successfully applied to this area, but current approaches require the use of large, painstakingly annotated datasets as training data~\cite{sigtia2016end} --- more often than not, the extensive MAPS piano database~\cite{maps}.
    However, curating additional training data can be both time-consuming and challenging~\cite{benetos2013automatic}, and the original setup used to create these datasets cannot be accurately reproduced should additional samples be needed.
    In addition, many of these machine learning approaches are both trained and evaluated on samples drawn from the same database~\cite{bock2012polyphonic, vd2009note, poliner2007discriminative, sigtia2016end}, weakening claims about generalization behavior; networks trained on one dataset tend to overfit its specific timbre and perform relatively poorly on newly generated data~\cite{poliner2007discriminative}.
    It has even been recently noted that neural networks face a fundamental issue when applied to polyphonic note transcription --- they suffer from the \textit{entanglement} problem, memorizing chords or combinations of notes rather than learning to report the onset of each note individually~\cite{entanglement}.
    
    We circumvent all of these problems by generating our training data procedurally.
    No annotated piano database is used as training data.
    Although we evaluate our approach on the MAPS piano database, the instruments and recording conditions used for testing are completely unknown to the network, providing a high degree of confidence in our model's generalization capabilities.
    Our approach is completely context-independent --- that is, we require no prior information about the instrument or recording being transcribed, allowing a broader field of application.
    Furthermore, we solve the disentanglement issue presented in \cite{entanglement} by randomly generating arbitrary combinations of notes, forcing the network to learn to identify individual notes.
    
    \section{Proposed Model}
    \subsection{Data Representation}
    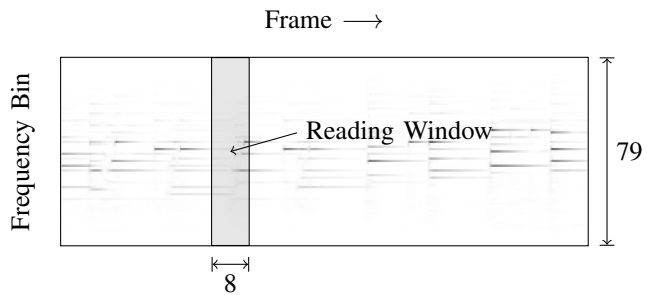
\begin{figure}
    	\centering
        \begin{tikzpicture}
        	\node[image, fill opacity=0.6] (spectrogram) at (3.5, 1.25)
    {\includegraphics[width=7cm]{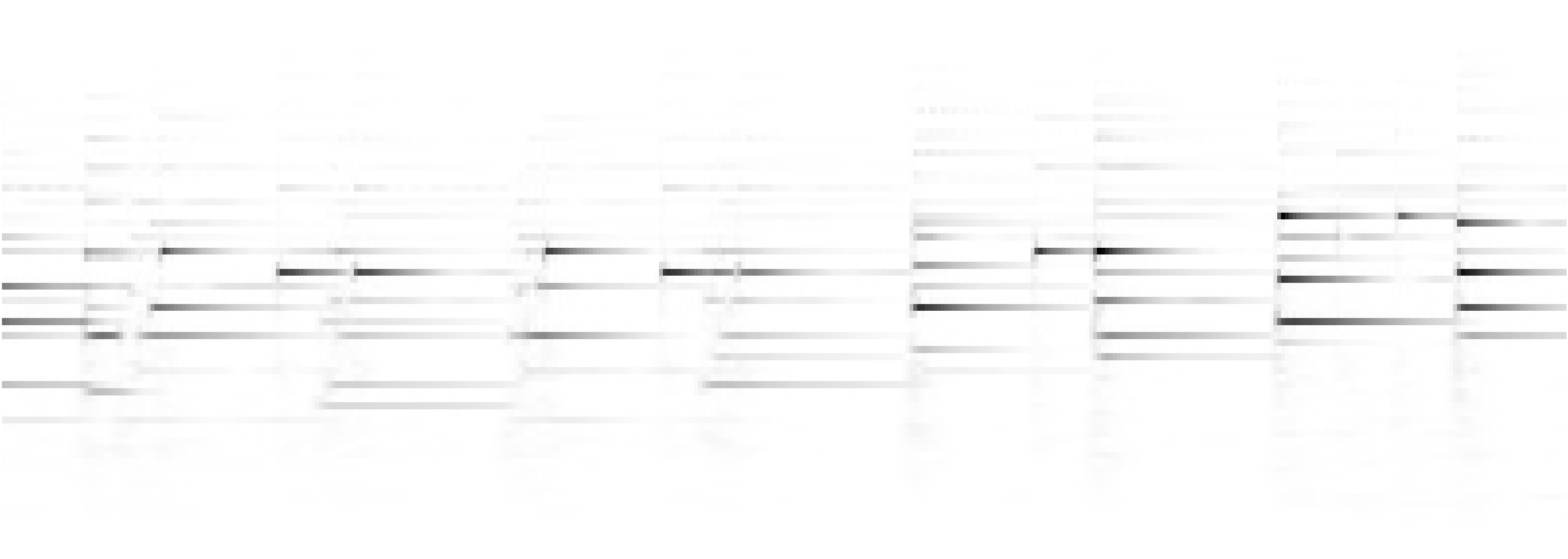}};
    		\draw (spectrogram.south west) rectangle (spectrogram.north east);
    
            \filldraw[fill=black, fill opacity=0.1] (2, 0) rectangle (2.5, 2.5);
            
            \node at ($(spectrogram.west) + (-0.5, 0)$) [rotate=90] {Frequency Bin};
            \node at ($(spectrogram.north) + (0, 0.5)$) {Frame $\longrightarrow$};
            \node (readingwindow) at (4.5, 1.5) {Reading Window};
            
            \draw[|<->|] ($(spectrogram.south east) + (0.25, 0)$) -- ($(spectrogram.north east) + (0.25, 0)$) node [midway, right] {79};
            \draw[|<->|] (2, -0.25) -- (2.5, -0.25) node [midway, below] {8};
            \draw[->] (readingwindow.west) -- (2.25, 1.25);     
        \end{tikzpicture}
        \caption{Visualization of reading window.}
        \label{fig:readingwindow}
    \end{figure}
    We use a constant-$Q$ transform (CQT) spectrogram as the fundamental time-frequency representation of our audio signals~\cite{cqt}.
    We use bins ranging from the note $\text{G}_1 \approx \SI{49}{Hz}$ to $\text{C}_8 \approx \SI{4186}{Hz}$, with a spacing of $12$ bins per octave, for a total of $79$ frequency bins.
    We use a $Q$-factor of $32$.
    
    Our spectrogram frames are spaced $1024$ audio samples apart; for the audio in the MAPS database, which has a sample rate of $\SI{44100}{Hz}$, this leads to a frame rate of about $\SI{43}{Hz}$.
    
    For our machine learning model, we use a simple feed-forward neural network.
    The network's input consists of an $8$-frame wide ``reading window'' of the magnitude of the CQT spectrogram, normalized to have a maximum value of $1$ (see Figure~\ref{fig:readingwindow}), yielding a total of $8 \times 79 = 632$ input values.
    The network's output is an $88$-dimensional vector; each component corresponds to a specific piano key, and represents the presence or absence of a note onset at the \nth{5} frame in the reading window.
    The output layer uses the sigmoid activation function, yielding values in the interval $(0, 1)$.
    We use two hidden layers of $512$ neurons each with the softsign activation function~\cite{softsign}.
    
    \subsection{Data Generation}
    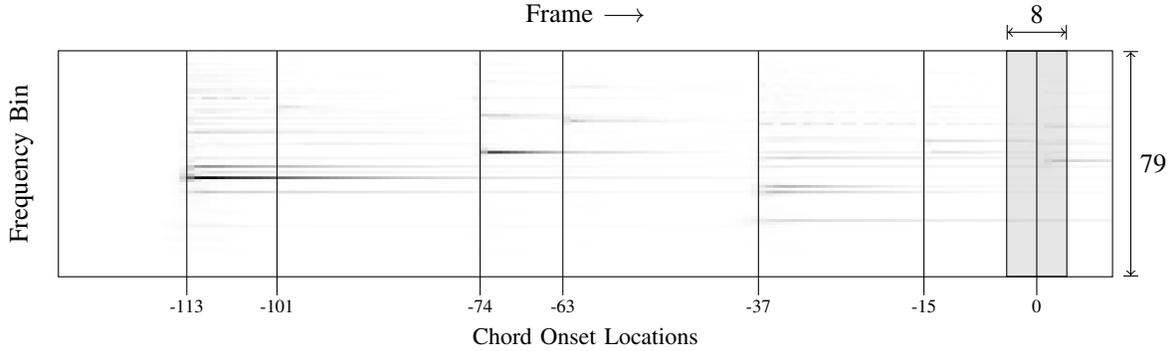
\begin{figure*}
    	\centering
        \begin{tikzpicture}[x=1mm, y=1cm]
        	\node[image] (spectrogram) at (-60, 1.5)
    {\includegraphics[width=14cm]{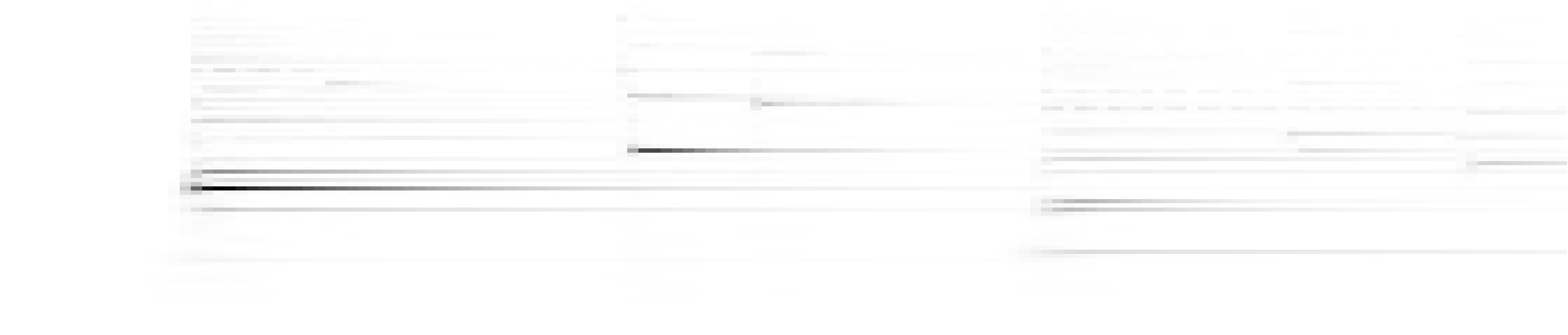}};
    		\draw (spectrogram.south west) rectangle (spectrogram.north east);
        
            \filldraw[fill=black, fill opacity=0.1] (-4, 0) rectangle (4, 3);
            
            \node at (-135, 1.5) [rotate=90] {Frequency Bin};
            \node at (-60, 3.5) {Frame $\longrightarrow$};
            \node at (-60, -0.8) {\small Chord Onset Locations};
            
            \draw[|<->|] (12.5, 0) -- (12.5, 3) node [midway, right] {79};
            \draw[|<->|] (-4, 3.25) -- (4, 3.25) node [midway, above] {8};
            
            \foreach \onset in {-113, -101, -74, -63, -37, -15, 0} {
            	\draw[thin] (\onset, -0.25) -- (\onset, 3);
                \node at (\onset, -0.4) {\scriptsize \onset};
            }
        \end{tikzpicture}
        \caption{Visualization of data generation algorithm (only spectrogram magnitude shown).}
        \label{fig:data-gen}
    \end{figure*}
    \begin{figure}
    	\centering
        \begin{tikzpicture}
        	\node[image] (window) at (1.5, 2)
    {\includegraphics[height=4cm]{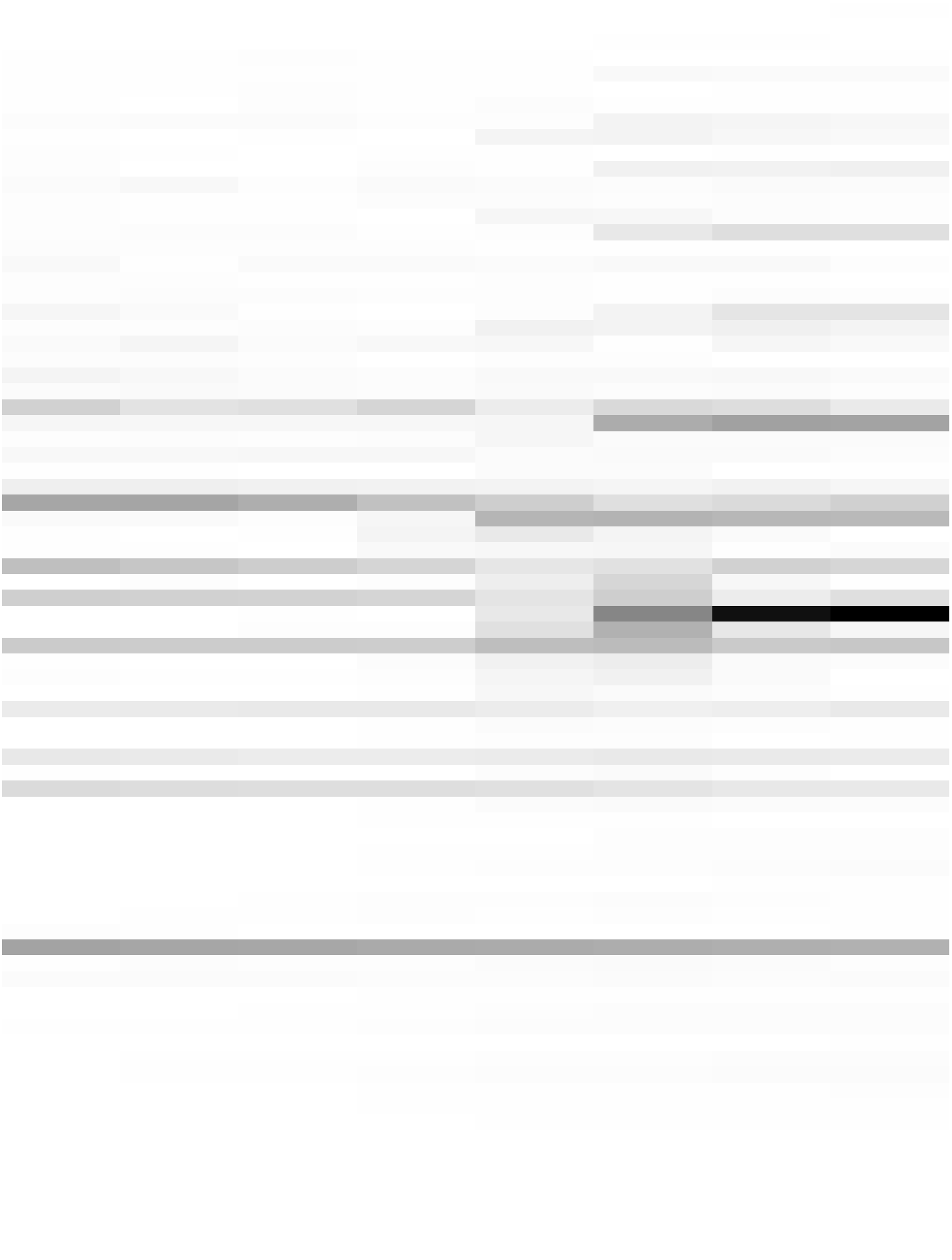}};
    		\draw (window.south west) rectangle (window.north east);
            
            \node[image] (label) at (5.5, 2)
    {\includegraphics[height=4cm]{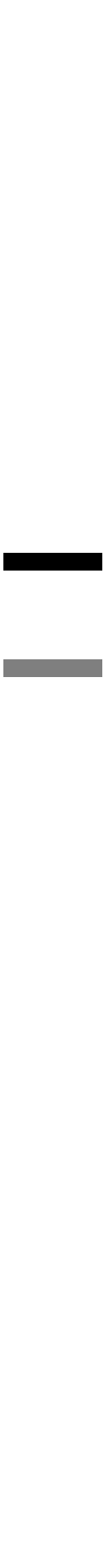}};
    		\draw (label.south west) rectangle (label.north east);
            
            \node at ($(window.west) + (-0.5, 0)$) [rotate=90] {Frequency Bin};
            \node at ($(window.north) + (0, 0.5)$) {Frame};
            \node at ($(window.south) + (0, -1)$) {\textsc{Input}};
            
            \node at ($(label.south) + (0, -1)$) {\textsc{Label}};
            
            \draw[|<->|] (3.25, 0) -- (3.25, 4) node [midway, right] {79};
            \draw[|<->|] ($(window.south west) + (0, -0.25)$) -- ($(window.south east) + (0, -0.25)$) node [midway, below] {8};
            \draw[|<->|] ($(label.south east) + (0.25, 0)$) -- ($(label.north east) + (0.25, 0)$) node [midway, right] {88};
            
            \foreach \note/\marking in {56/$\true$, 50/$\unknown$} {
            	\node at ($(label.south west) + (-0.5, {4 * \note/88})$) {\footnotesize \note: \marking};
            }
        \end{tikzpicture}
        \caption{Training example extracted from Figure~\ref{fig:data-gen}.}
    \end{figure}
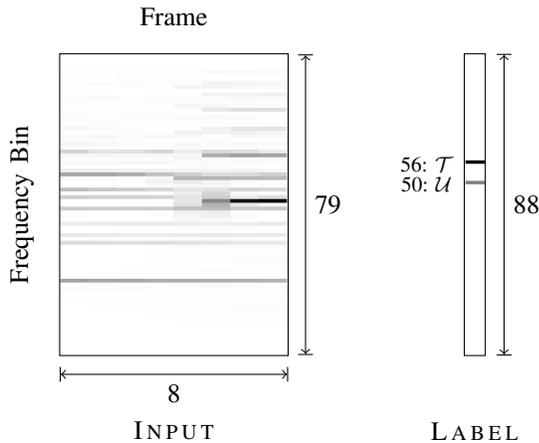
    Our data generation method is based upon the linearity of the CQT transform.
    By superimposing spectrograms of individual notes at various locations and intensities, we can model many characteristics of real piano music.
    
    For each training example, we generate an $8 \times 79$ magnitude spectrogram, representing the reading window, and its associated label.
    Each spectrogram is generated using a spectral basis: a collection of $88$ spectrograms, one for each piano note, each with a labeled onset time.
    
    For each of the MIDI instruments listed in Appendix~\ref{app:soundfont}, we generate a spectral basis for each of the four MIDI velocities $30$, $60$, $90$, and $120$.
    Spectrograms are generated from audio recordings of each note approximately $3.5$ seconds in length.\footnote{For SoundFont instruments, we use $0.5$ seconds of silence followed by $3$ seconds of sustained sound, leading to a $151$-frame spectrogram. Spectrograms for SFZ instruments are $172$ frames long due to a quirk in rendering, but this does not significantly affect our data generation.}
	This yields a total of $7 \times 4 = 28$ sets of spectral bases.
	Since loudly and softly played piano notes differ fundamentally in their harmonic composition, generating a new spectral basis for various MIDI velocities allows us to model the nonlinearities in note partials and decay associated with each dynamic level.
    
    Note that although the CQT transform is linear, its $magnitude$ is not.\footnote{However, the magnitude of the transform is $approximately$ linear, justifying NMF-based techniques~\cite{nmf}. Despite this, we find that many characteristics of real data, such as nonlinear note partial interference, are more effectively modeled by keeping complex phase information.}
    For this reason, we opt to store the complex-valued spectral basis and perform all calculations in the complex domain.
    
    \subsection{Algorithm Description}
    At the start of the generation of each training example, one of the spectral bases is randomly chosen.
    
    We label the \nth{5} frame in the reading window as Frame~0.
    Several chord onset times are randomly picked between Frame~$-130$ and Frame~$10$; the number of onset times is chosen from a Poisson distribution with~$\lambda = 6$.
    
    For each chord, we randomly pick several notes uniformly from $\text{A}_0$ to $\text{C}_8$ without replacement.
    The number of notes in each chord is chosen out of a geometric distribution with~$p = 0.4$.
    The spectrogram for the chord is modeled as a superposition of the spectrograms for each note.
    
    To improve robustness, we scale the basis spectrogram for each note by a factor randomly chosen between $0.1$ and $1$, and randomly shift the overall complex phase.
    The spectrogram of each note is also temporally shifted by a number of frames drawn from a discrete Gaussian distribution with $\sigma = 0.5$.
    
    To model note offsets, the spectrogram for a randomly chosen $5\%$ of notes begins to decay exponentially by a factor of $e$ per frame, starting on a frame randomly chosen from the reading window.
    
    Finally, the spectrogram for each chord is scaled by a factor randomly chosen between $0.1$ and $1$ and placed at its corresponding chord onset frame; the overall spectrogram is formed by superimposing all the chord spectrograms.
    
    We take the magnitude of the overall spectrogram\footnote{If no value within the reading window has a magnitude exceeding $10^{-3}$, we discard the data and start over to avoid ``silent'' frames.} and add white noise of magnitude $0.003$.
    We then use the reading window from Frame~$-4$ to Frame~$3$, normalized to a maximum value of 1, as our final generated data sample.\footnote{In practice, of course, we perform only the calculations necessary to compute the values within the reading window.}
    
    The labels are $88$-dimensional vectors whose components are ternary logical values.
    We assign a value of true ($\true$) if a note onset occurs at exactly the \nth{5} frame of the reading window, a value of unknown ($\unknown$) if an onset occurs at frames $3$, $4$, $6$, or $7$ of the reading window, and a value of ($\false$) otherwise.
    In practice, we encode the values $\true$, $\unknown$, and $\false$ as $1$, $\frac{1}{2}$, and $0$, respectively.
    
    \section{Training}
    \subsection{Setup}
    Training data was generated using the $24$ spectral bases associated with the \salamander, \fluid, \general, \kawai, \piano, and \city{} MIDI instruments.
    The network was trained for $1.5$ million iterations using the ADAM optimizer~\cite{adam} with an initial learning rate of $10^{-3}$.
    We use a mini-batch size of $32$ and an $L^2$ regularization parameter of $5 \times 10^{-10}$.
    
    We use a modified version of the cross-entropy loss function, where $\unknown$ labels do not contribute to the cost regardless of the network's output.
    This is to avoid penalizing ambiguous cases where a note onset occurs, but is not perfectly centered in the reading frame --- it is difficult to define a precise temporal threshold between the presence and absence of an onset.
    
    Detailed information about the system configuration used for training is available in Appendix~\ref{app:hardware}.
    
    \subsection{Validation}
    Three validation datasets were used, each of which contained $32768$ samples:
    \begin{enumerate}
    	\item Data generated using the $4$ spectral bases associated with the \shrine{} MIDI instrument.
        \item Synthesized piano recordings from the MAPS database.
        \item Real piano recordings from the MAPS database.
    \end{enumerate}
    
    For MAPS validation data, we precomputed the spectrogram for each piece and normalized each to a maximum value of 1.
    We randomly selected 1000 recordings for validation.
    Each sample was generated by selecting a reading window randomly from the validation spectrograms.
    Reading windows whose magnitudes did not exceed $10^{-3}$ were discarded; otherwise, they were normalized to a maximum of 1 and used as validation samples.
    Corresponding ternary labels were constructed using the MAPS onset annotations in the same manner as for synthesized data.
    Network performance on these validation datasets, as well as on a similarly generated dataset of $32768$ training samples, is discussed in Appendix~\ref{app:validation}.
    
    \section{Evaluation}
    Our network was evaluated on the $28910$ MAPS recordings not used for validation.
    For each spectrogram, we iterated over all positions of the reading window; windows were filtered and normalized as for the validation datasets, then fed through the network, generating a raw onset piano-roll representation.
    
    The onset piano-roll was thresholded at $0.8$ to obtain a binary activation piano-roll.
    For each of the $88$ notes, every run of consecutive activations was considered to be a single note onset event; the frame at which each event occurred was taken to be the average of all frames involved in a run.
    We convert the frame of each event into an onset time and evaluate our network predictions using the \texttt{mir\_eval} tool~\cite{mireval}.
    
    Onsets are considered to be correct if their onset time is within $\SI{50}{ms}$ of a ground truth onset at the same pitch; each label can only be matched with one network prediction.
    Unmatched labels are counted as false negatives, and unmatched predictions are counted as false positives.
    Standard note-based precision ($\precision$), recall ($\recall$), accuracy ($\accuracy$), and \mbox{f-measure} ($\fmeasure$) metrics are computed as described in~\cite{sigtia2016end}.

	\section{Results}
    \begin{table}
    	\caption{Evaluation Metrics on the MAPS Dataset}
        \centering
    	\begin{tabular}{lllll}
            \toprule
            Approach & $\precision$ & $\recall$ & $\accuracy$ & $\fmeasure$ \\
            \midrule
            Proposed & 85.46 & 59.79 & 54.27 & 70.36 \\
            Sigtia~\cite{sigtia2016end} & 67.75 & 66.36 & 50.07 & 67.05 \\
            B\"ock~\cite{bock2012polyphonic} & 93.3 & 91.5 & 85.9 & 92.4 \\
            Poliner~\cite{poliner2007discriminative} & --- & --- & 62.3 & ---\\
            Boogart~\cite{vd2009note} & --- & --- & 87.4 & --- \\
            \bottomrule
        \end{tabular}
        \label{tab:results}
    \end{table}
    \begin{table*}
    	\caption{Performance by MAPS Component (Disklavier~Only/Synthesized~Only/Overall)}
        \centering
    	\begin{tabular}{lllll}
            \toprule
            Component & $\precision$ & $\recall$ & $\accuracy$ & $\fmeasure$ \\
            \midrule
            Full & 79.13/87.46/85.46 & 55.16/61.27/59.79 & 48.16/56.32/54.27 & 65.01/72.06/70.36 \\
            MUS & 83.58/91.51/89.63 & 54.68/60.76/59.30 & 49.37/57.52/55.50 & 66.11/73.03/71.38 \\
            ISOL & 61.02/69.94/67.75 & 54.57/63.89/61.57 & 40.46/50.12/47.62 & 57.61/66.78/64.52 \\
            RAND & 71.38/82.72/79.76 & 57.13/62.52/61.17 & 46.48/55.29/52.95 & 63.47/71.21/69.24 \\
            UCHO & 68.16/74.20/72.65 & 59.67/63.37/62.44 & 46.66/51.92/50.56 & 63.63/68.35/67.16 \\
            \bottomrule
        \end{tabular}
        \label{tab:detailedresults}
    \end{table*}
    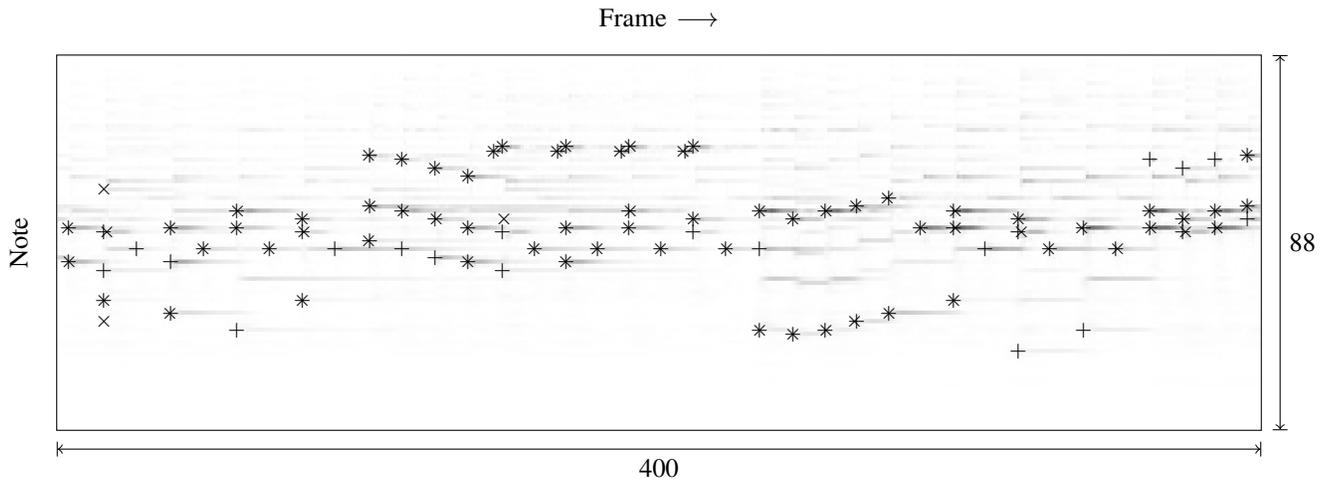
\begin{figure*}
    	\centering
        \newcommand{\windowwidth}{400}
        \newcommand{\imagewidth}{16}
        \newcommand{\imageheight}{5}
        \newcommand{\drawonset}[3]{\node at ($({\imagewidth * #2 / \windowwidth}, {\imageheight * (#1 - 0.5)/88})$) {\footnotesize #3};}
        \begin{tikzpicture}
        	\node[image, fill opacity=0.6] (spectrogram) at (8, 2.5)
    {\includegraphics[width=16cm]{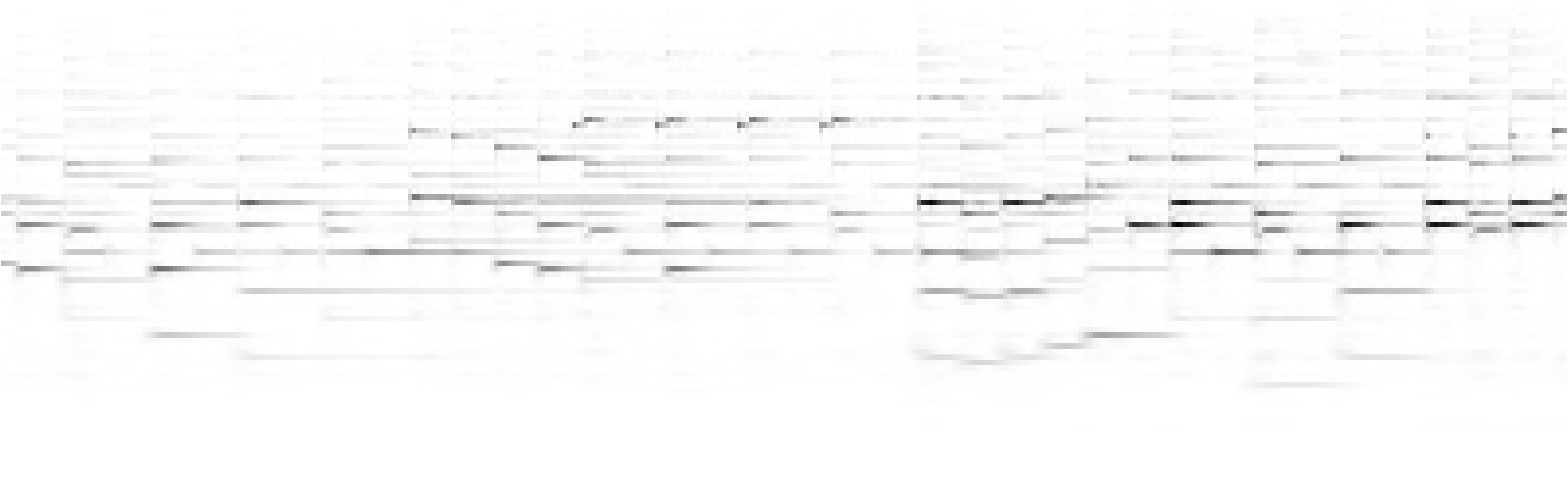}};
    		\draw (spectrogram.south west) rectangle (spectrogram.north east);
            
            \node at ($(spectrogram.west) + (-0.5, 0)$) [rotate=90] {Note};
            \node at ($(spectrogram.north) + (0, 0.5)$) {Frame $\longrightarrow$};
            
            \draw[|<->|] ($(spectrogram.south east) + (0.25, 0)$) -- ($(spectrogram.north east) + (0.25, 0)$) node [midway, right] {88};
            \draw[|<->|] ($(spectrogram.south west) + (0, -0.25)$) -- ($(spectrogram.south east) + (0, -0.25)$) node [midway, below] {\windowwidth};
            
            \foreach \pitch/\frame in {23/244.5, 24/233.5, 24/255.5, 26/15.5, 26/265.0, 28/37.5, 28/276.5, 31/15.5, 31/81.5, 31/298.0, 40/4.0, 40/136.5, 40/169.0, 43/48.5, 43/70.5, 43/158.5, 43/179.5, 43/200.5, 43/222.5, 43/329.5, 43/352.5, 45/103.5, 47/16.5, 47/82.0, 47/320.5, 47/375.0, 48/3.5, 48/37.5, 48/59.5, 48/136.5, 48/169.0, 48/190.0, 48/287.0, 48/298.5, 48/340.5, 48/363.5, 48/385.5, 50/81.5, 50/126.0, 50/148.5, 50/211.5, 50/244.5, 50/319.5, 50/374.5, 52/60.0, 52/115.0, 52/190.5, 52/233.5, 52/255.5, 52/298.5, 52/363.0, 52/385.0, 53/104.0, 53/265.5, 53/395.5, 55/276.0, 57/15.5, 60/136.5, 62/125.5, 64/114.5, 65/103.5, 65/395.5, 66/145.0, 66/166.0, 66/187.0, 66/208.0, 67/148.0, 67/169.0, 67/190.5, 67/211.5} {
                \drawonset{\pitch}{\frame}{$\times$};
            }
            
            \foreach \pitch/\frame in {19/319.4, 23/244.5, 24/59.7, 24/233.4, 24/255.2, 24/341.1, 26/265.7, 28/37.7, 28/276.4, 31/15.4, 31/81.5, 31/297.9, 38/15.4, 38/148.0, 40/3.7, 40/37.7, 40/136.5, 40/169.1, 41/125.6, 43/26.4, 43/48.6, 43/70.6, 43/92.3, 43/114.6, 43/158.6, 43/179.6, 43/200.6, 43/222.1, 43/233.4, 43/308.4, 43/329.9, 43/351.8, 45/103.9, 47/15.4, 47/81.5, 47/148.0, 47/211.3, 47/319.4, 47/374.1, 48/3.7, 48/37.7, 48/59.7, 48/136.5, 48/169.1, 48/190.0, 48/286.9, 48/297.9, 48/341.1, 48/363.1, 48/384.8, 50/81.5, 50/125.6, 50/211.3, 50/244.5, 50/319.4, 50/374.1, 50/395.6, 52/59.7, 52/114.6, 52/190.0, 52/233.4, 52/255.2, 52/297.9, 52/363.1, 52/384.8, 53/103.9, 53/265.7, 53/395.6, 55/276.4, 60/136.5, 62/125.6, 62/374.1, 64/114.6, 64/363.1, 64/384.8, 65/103.9, 65/395.6, 66/145.1, 66/166.4, 66/187.5, 66/208.7, 67/148.0, 67/169.1, 67/190.0, 67/211.3} {
                \drawonset{\pitch}{\frame}{$+$};
            }
        \end{tikzpicture}
        \caption{Our network predictions plotted against ground truth onset labels. Network predictions are shown with diagonal crosses; annotated onsets are shown with vertical crosses. The corresponding spectrogram has been placed to line up with ground truth onsets. The excerpt is a $400$-frame sample from the second movement of Mozart's Turkish March.}
        \label{fig:predictions}
    \end{figure*}
    
    Our results, as well as those of other machine learning systems, are shown in Table~\ref{tab:results}.\footnote{Poliner used a detection tolerance of $\SI{100}{ms}$ rather than $\SI{50}{ms}$.}
    Note that all other approaches listed use the MAPS dataset as both training and testing data, while we did not train on the MAPS dataset.
    In addition, our feed-forward neural network is much simpler than the models proposed by other authors.
    
    Nevertheless, we manage to exceed the \mbox{f-measure}, precision, and accuracy of the convolutional neural network used in~\cite{sigtia2016end} when evaluated on the MAPS dataset.
    Although~\cite{sigtia2016end} also gives results for a network trained on synthesized pianos and tested on real recordings, their results under this configuration are significantly worse: training and testing on different datasets, they achieve an onset \mbox{f-measure} of only $54.89$.
    
    Since we do not train on MAPS data, our results demonstrate our system's ability to transcribe recordings under novel conditions.
    Our approach is also context-independent, allowing it to be immediately applied to any recording without prior adjustment or fine-tuning.
    
    Detailed results for the overall MAPS dataset, as well as for each individual component, are listed in Table~\ref{tab:detailedresults}.
    Note that our network achieves the best \mbox{f-measure} on the MUS component, which consists of real musical compositions.
    A sample comparison between our network's predictions and the MAPS ground truth labels is shown in Figure~\ref{fig:predictions}.
    
    While our network performs significantly worse on the real Disklavier samples, these recordings can have annotation discrepancies of up to \SI{100}{ms}~\cite{ewert2016piano}.
    In addition, we found several issues with the MAPS Disklavier dataset, including omitted notes and recordings consisting entirely of percussive keybed and pedal noises.
    However, even taking these discrepancies into account, we do expect our network to perform worse on real piano recordings; software samples are unable to capture the piano's full range of acoustic subtleties.

    \section{Conclusion}
    We have presented a method of generating spectral data which accurately models many aspects of piano music, avoiding the need for an annotated musical database.
    We trained a simple machine learning model on our procedurally generated data and achieve good transcription results under novel recording conditions, highlighting our network's context-independence and excellent generalization qualities.
    Our approach outperforms that of~\cite{sigtia2016end}, even though they train and test on samples drawn from the same database and propose a much more complex network architecture.
    
    \subsection{Future Work}
    Several improvements to our data generation method could be implemented:
    \begin{itemize}
    	\item We could quite easily model the full range of pianos more accurately by adding more MIDI instruments or increasing the number of MIDI velocities sampled.
    	\item Note offsets are modeled in a rather crude fashion by applying exponential decay to the note spectrogram; a sample-based approach could more accurately model various articulations.
    	\item No distinction is made between notes played with and without sustain pedal --- lifting the dampers alters note timbre slightly and could be modeled with additional samples.
    	\item Rather than model chord onset times and notes randomly, a musical model could be introduced to take into account the temporal patterns and harmonic progressions present in Western music.
    \end{itemize}
    
    In addition, our method could easily be extended to generate training data for recursive, convolutional, or LSTM networks.
    Considering that we were able to train a very simple network network using our proposed method and achieve good transcription performance, using a more complex machine learning model would likely yield better results.
    
    The use of generated training data as input to a neural network brings to mind the principle behind generative adversarial networks, which have not yet been applied to piano onset transcription.
    It may be possible to train a neural network to generate data which closely resemble those of real pianos, effectively creating an infinite stream of high-quality spectral data indistinguishable from those of an annotated database.

    \appendices
    \section{MIDI Instruments}
    \label{app:soundfont}
    \begin{table}
    	\caption{List of SoundFont/SFZ Instruments}
        \centering
        \begin{tabular}{lll}
            \toprule
            Abbreviation & Name & Type\\
            \midrule
            \salamander & Salamander Grand Piano & SFZ \\
            \fluid & Fluid R3 GM & SF2\\
            \general & GeneralUser GS & SF2\\
            \kawai & Kawai Upright Piano & SFZ\\
            \piano & Yamaha Disklavier Grand Piano & SF2 \\
            \city & The City Piano & SFZ\\
            \shrine & The Shrine Piano & SFZ \\
            \bottomrule
        \end{tabular}
        \label{tab:instruments}
    \end{table}
    The SoundFont/SFZ instruments used are listed in Table~\ref{tab:instruments}; all of them are available online, free of charge.
    Audio samples for spectral basis generation were synthesized by FluidSynth for SoundFont~2 instruments and by Plogue sforzando for SFZ instruments.
    Both software synthesizers are available free of charge.

    \section{Training and Validation Performance}
    \label{app:validation}
    \begin{figure}
    	\centering
        \newcommand{\labelshift}{-0.05}
        \newcommand{\lowlabel}{0.2}
        \newcommand{\labelsize}{\scriptsize}
        \begin{tikzpicture}[scale=0.7]
        	\draw[step=2] (0, 0) grid (6, 6);
            \node at (1, 6.4) {$\true$};
            \node at (-0.4, 1) {$\false$};
            \node at (3, 6.4) {$\unknown$};
            \node at (-0.4, 3) {$\unknown$};
            \node at (5, 6.4) {$\false$};
            \node at (-0.4, 5) {$\true$};
            \node at (3, 7) {Label};
            \node at (-1.1, 3) [rotate=90] {Prediction};
            
            \node (HFN) at (1, 1.4) {HFN};
            \node (STN) at (3, 1.4) {STN};
            \node (HTN) at (5, 1.4) {HTN};
            \node (SFN) at (1, 3.4) {SFN};
            \node (VC) at (3, 3.4) {VC};
            \node (SFP) at (5, 3.4) {SFP};
            \node (HTP) at (1, 5.4) {HTP};
            \node (STP) at (3, 5.4) {STP};
            \node (HFP) at (5, 5.4) {HFP};
            
            \node[below=\labelshift of HFN, align=center] {\labelsize Hard False};
            \node[below=\labelshift of STN, align=center] {\labelsize Soft True};
            \node[below=\labelshift of HTN, align=center] {\labelsize Hard True};
            \node[below=\labelshift of SFN, align=center] {\labelsize Soft False};
            \node[below=\labelshift of VC, align=center] {\labelsize Vacuously};
            \node[below=\labelshift of SFP, align=center] {\labelsize Soft False};
            \node[below=\labelshift of HTP, align=center] {\labelsize Hard True};
            \node[below=\labelshift of STP, align=center] {\labelsize Soft True};
            \node[below=\labelshift of HFP, align=center] {\labelsize Hard False};
            
            \node[below=\lowlabel of HFN, align=center] {\labelsize Negative};
            \node[below=\lowlabel of STN, align=center] {\labelsize Negative};
            \node[below=\lowlabel of HTN, align=center] {\labelsize Negative};
            \node[below=\lowlabel of SFN, align=center] {\labelsize Negative};
            \node[below=\lowlabel of VC, align=center] {\labelsize Correct};
            \node[below=\lowlabel of SFP, align=center] {\labelsize Positive};
            \node[below=\lowlabel of HTP, align=center] {\labelsize Positive};
            \node[below=\lowlabel of STP, align=center] {\labelsize Positive};
            \node[below=\lowlabel of HFP, align=center] {\labelsize Positive};
        \end{tikzpicture}
        \caption{Confusion matrix under ternary logic.}
        \label{fig:ternary}
    \end{figure}
    
    \begin{table}
    	\caption{Training and Validation Performance}
        \centering
    	\begin{tabular}{lllll}
            \toprule
            Dataset & $\precision$ & $\recall$ & $\accuracy$ & $\fmeasure$ \\
            \midrule
            Generated Training & 99.3 & 53.7 & 72.5 & 69.7 \\
            Generated Validation & 96.9 & 46.9 & 66.1 & 63.2 \\
            Synthesized MAPS & 90.8 & 47.2 & 67.6 & 62.1 \\
            Real MAPS & 81.2 & 32.7 & 54.2 & 46.7 \\
            \bottomrule
    	\end{tabular}
        \label{tab:validation}
    \end{table}
    
    For evaluation on training and validation datasets, the network's output is interpreted as a ternary logical value; a value greater than $0.8$ is reported as a note onset ($\true$), a value less than $0.2$ is reported as a lack thereof ($\false$), and a value of $\unknown$ is reported otherwise.
    We evaluate the network's precision~($\precision$), recall~($\recall$), accuracy~($\accuracy$), and \mbox{f-measure}~($\fmeasure$), defined using the ternary confusion matrix (Figure~\ref{fig:ternary}) as follows:
    
    \begin{align*}
    	\precision &= \frac{\text{HTP} + \text{STP}}{\text{HTP} + \text{STP} + \text{HFP}} \\
        \recall &= \frac{\text{HTP}}{\text{HTP} + \text{SFN} + \text{HFN}} \\
        \accuracy &= \frac{\text{HTP + STP}}{\text{HTP} + \text{STP} + \text{HFP} + \text{SFN} + \text{HFN}} \\
        \fmeasure &= \frac{2 \times \precision \times \recall}{\precision + \recall} \\
    \end{align*}

	Our network's performance on training and validation data can be found in Table~\ref{tab:validation}.
    Since these metrics are computed in a non-standard manner, they cannot be directly compared with those of other approaches.
    However, we observe that the \mbox{f-measures} for our generated validation data and the synthesized MAPS validation data are nearly identical, highlighting our network's generalization abilities.
    
    \section{Training Setup}
    \label{app:hardware}
    Our network was trained on the CPU of a Lenovo \mbox{Z50-70} laptop.
    The system has a \SI{1.40}{GHz} Intel Celeron 2957U CPU and \SI{4.00}{GB} of DDR3 RAM.
    We use TensorFlow for Python~3.5.3 on Windows~10.
    Training took a total of $44$ hours.

    \printbibliography
\end{document}